\documentclass{article}
\usepackage{spconf,amsmath,graphicx,hyperref, amsmath, amssymb}
\DeclareMathOperator*{\argmin}{argmin} 
\usepackage{breqn}
\usepackage{caption}
\usepackage{subcaption}

\title{Shared Neural Space: Unified Precomputed Feature Encoding for Multi-Task and Cross-Domain Vision}
\name{Jing Li, Oskar Bartosz\sthanks{Samsung Research Poland}, Chengyu Wang, Michal Wnuczynski\sthanks{Samsung Research Poland}, Dilshan Godaliyadda, Michael Polley}
\address{MPI Lab\\Samsung Research America\\ 6105 Tennyson Parkway, TX, USA}
\begin{document}
%
\maketitle
\begin{abstract}
The majority of AI models in imaging and vision are customized to perform one specific high-precision task.
However, this strategy is inefficient for applications with a series of modular tasks, since each requires a mapping to a disparate latent domain. 
To address this inefficiency, we propose a universal Neural Space (NS), where an encoder–decoder framework pre-computes features across vision and imaging tasks. 
Our encoder learns transformation-aware, generalizable representations, which enable multiple downstream AI modules to share the same feature space. 
This architecture reduces redundancy, improves generalization across domain shifts, and establishes a foundation for efficient multi-task vision pipelines.
Furthermore, as opposed to larger transformer backbones, our backbone is lightweight and CNN-based, allowing for more wider use across hardware. 
We further demonstrate that imaging and vision modules such as demosaicing, denoising, depth estimation and semantic segmentation can be performed efficiently in this NS. 
\end{abstract}
%
\section{Introduction}
\label{sec:intro}

Modern vision/imaging systems commonly train separate task-specific AI models, each with a disparate latent space. 
While this approach is well suited for free-standing AI modules, it is highly inefficient for applications with multiple parallel or sequential AI modules, since mapping back and forth from different latent domains result in compounded inefficiency.

\begin{figure}[t]
\centerline{\includegraphics[width=0.9\columnwidth]{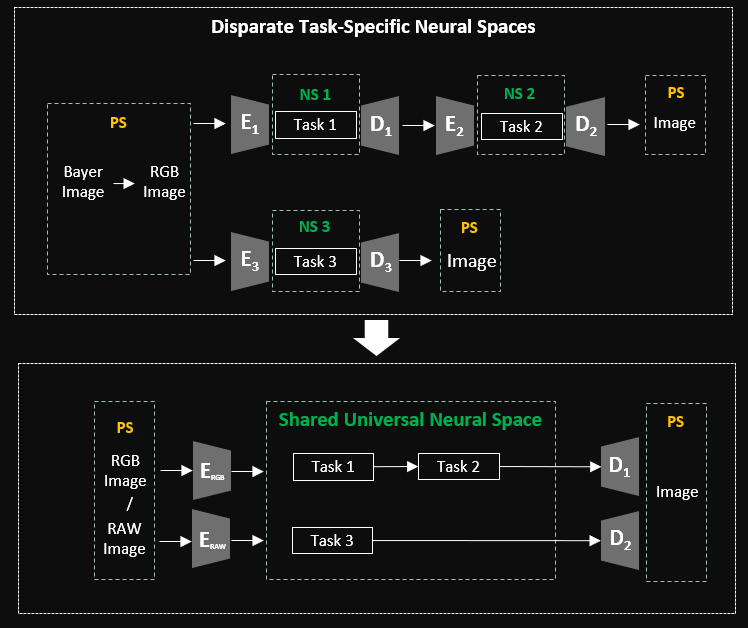}}
\caption{An overview of our shared NS framework: 
Top: How vision systems operate today -- disparate NS per task. Bottom: proposed NS -- multiple tasks performed in NS. 
}
\label{fig:overview}
\end{figure}

In this paper, we propose a CNN-based encoder--decoder framework that learns a
\emph{Shared Neural Space (NS)} reusable across diverse vision/imaging tasks. 
The encoder is pretrained on large-scale data, so that task-specific modules can consume its
features instead of raw pixels. This design reduces redundancy, thereby making
multi-task pipelines more efficient.
The core idea behind the NS is illustrated in Fig.~\ref{fig:overview} -- i.e. the efficiency inherent in a shared NS for imaging as well as vision tasks. 
Our NS is \emph{transformation-aware} i.e. supports affine transforms, thereby making it a technical analogue of an “intermediate image format” tailored specifically for AI applications, enabling direct reuse across modular tasks.
Furthermore, thanks to its CNN backbone, we have been able to export the encoder to mobile/embedded environments using standard deployment tool-chains.

To demonstrate the utility of the NS, we focus on three canonical, complementary, and representative dense prediction tasks that require rich spatial context: \emph{semantic segmentation} which is categorical and spatially discrete, \emph{depth estimation} which is continuous and geometric and \emph{denoising} which is spatially continuous. Experiments demonstrate that models built on the Shared NS
achieve strong generalization across tasks and domains. 
In particular, when evaluated under distribution shifts such as training on synthetic
data and testing on real images, our approach consistently outperforms
pixel-based alternatives. 
These results highlight both the robustness and the practical re-usability of the Shared NS as a foundation for
vision tasks. 
Our specific contributions are:
\begin{itemize}
  \item A CNN-based encoder--decoder framework yielding a Shared Neural
        Space across vision tasks, with verified mobile deployment.
  \item Demonstration that the representation supports structured image
        operations, facilitating feature reuse.
  \item Empirical evidence that the approach improves generalization under domain shifts, trading minor in-distribution loss for stronger robustness.
\end{itemize}

\section{Related Work}
\label{sec:relwork}

Research on general-purpose feature learning has steadily progressed from
early CNN activations reused across tasks~\cite{donahue2014decaf,razavian2014offtheshelf}
to large-scale self-supervised and multimodal training
\cite{chen2020simclr,he2022mae,radford2021clip}.
These approaches demonstrated that features learned once can serve a wide
variety of downstream objectives. In parallel, multi-task learning has
established that shared representations can benefit diverse problems,
whether through joint optimization~\cite{caruana1997mtl,misra2016crossstitch,kendall2018uncertainty}
or more structured transfer analyses such as Taskonomy~\cite{zamir2018taskonomy}.
Universal backbones trained across many objectives
\cite{kokkinos2017ubernet,vandenhende2020mtinet} further reinforced the
value of decoupling low-level feature extraction from task-specific heads.

Another line of work emphasizes robustness across domains, where models are
expected to handle distribution shifts such as synthetic-to-real transfer. Adaptation and augmentation methods~\cite{tsai2018adaptsegnet,hoffman2018cycada,zhou2021mixstyle,cha2021swad} propose ways to bridge visual gaps. These results collectively suggest that
representations capturing structure beyond raw pixel statistics are more
likely to generalize across shifts . Our approach focuses on learning a representation that inherently reduces cross-domain divergence, thereby improving transfer without the need for explicit adaptation mechanisms.



  
\section{Methodology}
\label{sec:method}



\subsection{Encoder--Decoder Design}
\label{sec:Encoder--Decoder}
In this section, we discuss how we design a universal NS for different image formats such as RGB and RAW. Specifically, we will design two encoders, to map from RGB/RAW$\rightarrow$NS, and a decoder to map from NS$\rightarrow$RGB. 
Hence, the input to the encoder can be $X_{RGB} \in \mathbb{R}^{3\times H \times W}$ for an RGB image or ${X_{RAW}} \in \mathbb{R}^{1\times H \times W}$ for a RAW image. 
We then define the features of the NS as, 
\begin{equation} z_t, z_b = E(X)
\end{equation}
where, $z_b \in \mathbb{R}^{64 \times \frac{H}{2} \times \frac{W}{2}},  z_t \in \mathbb{R}^{64 \times \frac{H}{4} \times \frac{W}{4}}, z=[z_t, z_b] $
for a given encoder $E(\cdot)$, and an input $X$ 

In our framework, we first train a RGB$\rightarrow$NS encoder, together with a NS$\rightarrow$RGB decoder, by using the same RGB image as input and output.   
Here, in order to ensure that the NS is equivariant under transforms, we train the network to produce the same result under pixel space (PS) and NS transforms. 
Then, we freeze the NS$\rightarrow$RGB decoder,and train a RAW$\rightarrow$NS encoder, using corresponding RAW and RGB images.
It is important to note that for the encoder-decoder, we used the architecture in \cite{Li_2025_CVPR} without vector quantization.

\subsubsection{RGB$\rightarrow$NS Encoder and NS$\rightarrow$RGB Decoder Design}
\label{sec:rgb-nscc}

Here we elaborate further on designing the RGB$\rightarrow$NS Encoder, $E_0(\cdot)$, and the NS$\rightarrow$RGB decoder, $D_0(\cdot)$. 
Let the parameters of the encoder be $\theta_{e_0}$ and parameters of the decoder be $\theta_{d_0}$.
Then, our goal is to estimate, 
\begin{equation}
    \theta_{e_0}^*, \theta_{d_1}^* = \argmin_{\theta_{e_0},\theta_{d_1}}  \{ L_{rl}[X, D_0(E_0(X))] + L_{reg} (E_0, D_0, f) \}
\end{equation}
where $L_{rl}$ is the image reconstruction loss computed as the weighted sum of L1 and SSIM losses \cite{Li_2025_CVPR}, and $L_{reg}$ is a regularizer that ensures transform equivariance across pixel and latent domains. 
We implemented the regularizer as a loss function in both NS and the PS. The idea is similar to what was used in \cite{kouzelis2025eqvae}, but here it is purely for image representation, not generation. 
\begin{equation}
L_{reg}=L_2[z^f, E_0({X^f})] + L_{rl}[\hat{X^f}, D_0(z^f)]
\end{equation}
where $f$ is random affine transform such as rotation, scaling and translation. 


\subsubsection{RAW$\rightarrow$NS Encoder Design}
Next, to compute the parameters of the RAW$\rightarrow$NS encoder, $E_1(\cdot)$, we freeze the NS$\rightarrow$RGB decoder and train only $E_1(\cdot)$, using a RAW image as input, and the corresponding RGB image, obtained by demosaicing the RAW image using RSTCA \cite{xing2022residualswintransformerchannel}, as output of $D_1(\cdot)$. 
\begin{equation}
    \theta_{e_1}^* = \argmin_{ \theta_{e_1}} L(X_{RGB}, X_{RAW}, E_1, D_1^*)
\end{equation}
where $L(*)$ denotes a loss function, $D_1^*(\cdot)$ is the frozen decoder from sec \ref{sec:rgb-nscc} and $\theta_{e_1}$ is the parameters for $E_1(\cdot)$ .
The loss function is a combination of $L_2$ loss in NS and reconstruction loss in PS, which is defined as 
\begin{equation}
    L = L_2[E_1(X_{RAW}), z] + \\
    L_{rl}[D_1(E_1(X_{RAW})), D_1^*(z)]
\end{equation}
where $z$ is the NS feature for $X_{RAW}$.

\subsection{Neural Space Interface for Downstream Tasks}
In this section, we elaborate on how the downstream modular tasks are architected within the shared NS. 
In particular, we utilize the precomputed features directly in the downstream module, instead of recomputing features, which in turn allows for a more efficient implementation of said module. 



\subsubsection{Neural Space Interface for Denoising}
The NS features, $z_b$ and $z_t$, have different spatial dimensions. While the former has higher spatial dimensions and preserves more high frequency details of the image, the latter contains more structural information of the image. For denoising, we trained a denoiser, $F_{denoise}$, that only denoised $z_b$. The denoiser consists of six residual blocks, each containing two convolutional layers, and it is trained with the following loss function
\begin{equation}
\begin{aligned}
    L_{denoiser}=&L_2[F_{denoise}(E(X_{gt})),F_{denoise}(E(X))] \\
    +&\lambda L_{re}[X_{gt}, D_1(F_{denoise}(E(X)))],
\end{aligned}
\end{equation}
where ($X$,$X_{gt}$) is the paired input-GT training data, and $\lambda$ is a weighting scalar.
 
\subsubsection{Neural Space Interface for Semantic Segmentation}
For semantic segmentation, we integrate $z_b$ directly into a segmentation backbone by replacing the initial convolutional layer
with the encoder output. 
The remainder of the architecture is kept unchanged, enabling efficient reuse of precomputed features while retaining
the strong multi-branch design of the original model. 
The segmentation decoder, $D_{seg}$, is trained end-to-end with cross-entropy loss, as in Equation \ref{eq:seg_eq}, where $\ell_{\text{CE}}$ is the pixel-wise cross-entropy.
\begin{equation}
\label{eq:seg_eq}
\begin{aligned}
    L_{seg} = L_{\text{CE}}\big[D_{seg}(E(X)), \, Y\big],
\end{aligned}
\end{equation}
It is important to note that we refer to the segmentation module as a "decoder" since, it directly transforms from NS to labeled PS. 

\subsubsection{Neural Space Interface for Stereo Depth Estimation}
For stereo depth estimation, we adapt the backbone to integrate
the Shared NS encoder outputs into a multi-scale feature
hierarchy. 
The encoder produces dual latent streams $(z_b, z_t)$, 
that are processed by a sequence of down-sampling blocks and
fused through FPN layers, yielding feature pyramids
$\{p_2, p_3, p_4, c_5\}$ at progressively smaller resolutions.

The depth decoder $D_{depth}$ predicts disparity, upscaling it with learned method. 
In the process of training we ensure that the lower resolution disparity map encodes useful low-frequency information by training on bilinearly upscaled maps. 
Training uses a multi-scale smooth L1 loss from Equation \ref{eq:depest}, where $d_{gt}$ is the ground-truth disparity and $\ell_{\text{SL1}}$ is the smooth L1 loss applied over valid pixels, and $U_{learned}$ and $U_{bilinear}$ are upscaling methods. We set $\alpha=0.3$ to balance the auxiliary term.
\begin{equation}
\label{eq:depest}
\begin{aligned}
    L_{depth} = & \; L_{\text{SL1}}\big[U_{learned}(D_{depth}(E(X))), \, Y\big] \\
                &+ \alpha \, L_{\text{SL1}}\big[U_{bilinear}(D_{depth}(E(X))), \, Y\big],
\end{aligned}
\end{equation}

\section{Experiments}
\label{sec:exp}
We evaluate the proposed framework across representative dense prediction
tasks, measuring its ability to generalize under distribution shifts,
handle different input domains, and reduce computational overhead. All
experiments were conducted on an NVIDIA A100 GPU and S25Ultra chipset. We report 
both quantitative results (tables and metrics) and qualitative evidence 
(visualizations) to highlight The robustness and efficiency of the Shared NS.

\subsection{Neural Space Training Strategy}
For the RGB encoder-decoder training, we use the same setup and dataset as in paper \cite{Li_2025_CVPR}.
For Bayer encoder-decoder, we first pass the Bayer image into RSTCA \cite{xing2022residualswintransformerchannel} to generate the corresponding RGB image. We use 500 image pairs to training the encoder for Bayer encoder.

\subsection{Performance on Different Input Domains}
A desirable property of the Shared Neural Space is that it produces
consistent representations when the same scene is captured in different
input formats. In particular, we examine RGB and RAW image domains, which
differ substantially in appearance but carry the same semantic content. An
effective encoder should map both into overlapping distributions. 



\begin{figure}
    \centering
    \begin{subfigure}[b]{0.23\textwidth} 
        \centering
        \includegraphics[width=\linewidth]{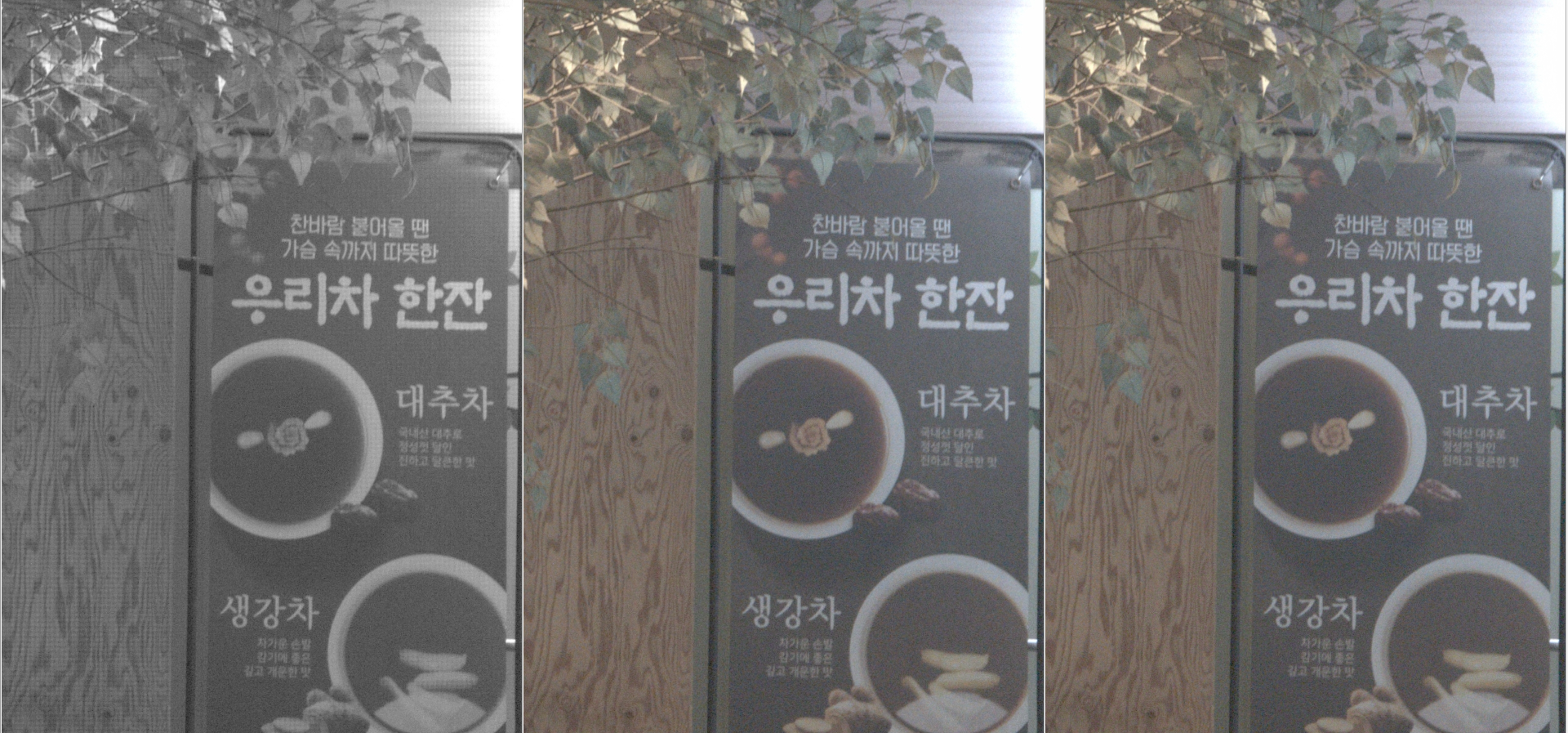} 
        \caption{RGB/RAW NS encodings}
        \label{fig:rgb_raw_latents}
    \end{subfigure}
    \hfill 
    \begin{subfigure}[b]{0.23\textwidth} 
        \centering
        \includegraphics[width=\linewidth]{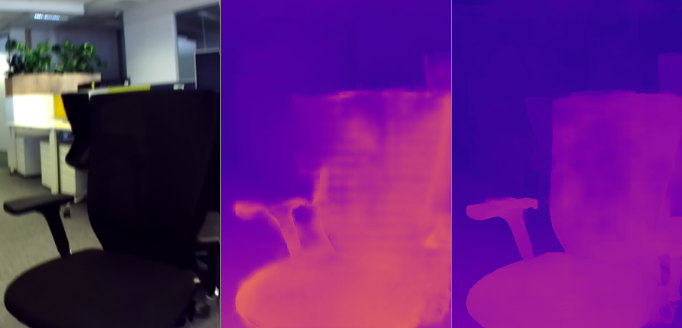} 
        \caption{Depth Generalization}
        \label{fig:gen_dep_est}
    \end{subfigure}
    \caption{(a). Comparison of latent encodings produced from corresponding RGB and RAW inputs of the same scene. From left: Bayer image, decoded bayer NS image, decoded RGB NS image. (b). Depth estimation generalization capabilities tested on unseen data distribution. From left: GT, PS model, encoder-based model.}
\end{figure}

Figure~\ref{fig:rgb_raw_latents} illustrates qualitative results, showing
pairs of RGB and RAW inputs from the same scene and their corresponding
latent encodings. 
As is evident, both RGB input and Bayer input are decoded into the same image,
which implies that they have shared neural space

\subsection{Cross-Domain Semantic Similarity}
In this section we investigate how the Shared NS preserves semantic consistency under visual shifts. 
The goal is to evaluate whether semantically equivalent images, even when altered in
appearance, remain closer in the NS induced by our encoder
compared to task-specific PS embeddings.

We construct the experiment as follows. A set of 30 Middlebury images is transformed with 8 distribution-shifting
functions that alter their appearance while preserving semantics, using imagenet-c (motion blur, jpeg, gausian noise, etc.). 
All images are then encoded with both (i) the Shared NS encoder $E(\cdot)$ and (ii) feature
embeddings extracted from task-specific PS models.

For each group, we compute the L1 norm between the
embeddings of the original set and those of its shifted
variants. Each metric is normalized by range of embeddings, and by number of features to compare. 
Shared NS, being transformation-aware, produces embeddings 
that remain closer across such shifts, while pixel-based feature 
extractors show larger discrepancies.

Results are summarized in Table~\ref{tab:semantic_similarity}, showing the
average L1 norm (lower is better) across all transformations.

\begin{table}[h]
\centering
\caption{Cross-domain semantic similarity measured
between embeddings of original and transformed images}
\label{tab:semantic_similarity}
\begin{tabular}{lcccc}
\hline
Type  & Avg L1 & Mean L1 \\
\hline
PS embeddings  &  0.018766  & 0.008821 \\
Shared Neural Space  &  \textbf{0.009419}  & \textbf{0.006948} \\
\hline
\end{tabular}
\end{table}
\vspace{-5mm}

\subsection{Cross-Domain Generalization Metrics}
We first evaluate cross-domain generalization, where models are trained on
one dataset and tested on another without adaptation. This setting
highlights the robustness of the Shared NS compared to
PS baselines. Tables \ref{tab:seg_camdvid} and \ref{tab:depth_cross} presents metrics obtained from testing on unseen data distributions. Visual results for depth generalization are shown in \ref{fig:gen_dep_est}.


To demonstrate the performance of the denoising model and show the benefits of the NS, we trained the proposed model as well as a Pixel space (PS) denoising counterpart using DnCNN~\cite{zhang2017beyond}. The number of trainable parameters and the runtime of the two denoiser, tested on $512\times512$ images, are summarized in Table~\ref{tab:denoise}. The runtime for the Neural Space (NS) denoiser only considers the denoiser itself. However, even if the runtime of the encoder/decoder is included, the NS denoising only takes 0.82s. 

\begin{table}[h]
\centering
\caption{Compare denoising performance.}
\label{tab:denoise}
\begin{tabular}{lccc}
\hline
Method & \# of parameters & runtime (s) \\
\hline
PS denoiser & 558336 & 1.46\\
NS denoiser & 443136 & 0.21\\
\hline
\end{tabular}
\end{table}


\begin{table}[h]
\centering
\caption{Cross-domain semantic segmentation performance on CamVid dataset (trained on Cityscapes).}
\label{tab:seg_camdvid}
\begin{tabular}{lccc}
\hline
Method & MeanIU $\uparrow$ & Pixel Acc. $\uparrow$ & Mean Acc. $\uparrow$ \\
\hline
PS & 0.5893 & 0.8709 & 0.7223 \\
NS & \textbf{0.6013} & \textbf{0.8725} & \textbf{0.7684} \\
\hline
\end{tabular}
\end{table}

\begin{table}[h]
\centering
\caption{Cross-domain depth estimation performance on Middlebury dataset (trained on Sceneflow)}
\label{tab:depth_cross}
\begin{tabular}{lcccc}
\hline
Method & d1 all $\downarrow$ & Treshold 3 $\downarrow$ \\
\hline
Pixel space (PS) & 14.9911 & 17.5367 \\
Neural Space (NS)  & \textbf{14.1881} & \textbf{16.9185} \\
\hline
\end{tabular}
\end{table}
\vspace{-5mm}

\subsection{Computational Cost}
Since features are precomputed once and reused, downstream
decoders operate on reduced feature maps instead of raw pixels. We evaluate
both inference time and computational complexity (GFLOPs) for semantic
segmentation, comparing encoder-based models with PS baselines.
Similar results are observed for depth estimation, we skip the results to save space.

\begin{table}[h!]
\centering
\caption{Encoder inference time and GFLOPs.}
\label{tab:comp_enc}
\begin{tabular}{lcccc}
\hline
Resolution & Time (s) & GFLOPs \\
\hline
1280$\times$720  & 0.015 & 95.55 \\
1920$\times$1080 & 0.033 & 214.99 \\
\hline
\end{tabular}
\end{table}

\begin{table}[h!]
\centering
\caption{Computational cost for semantic segmentation.}
\label{tab:comp_seg}
\begin{tabular}{lcccc}
\hline
Resolution & Method & Time (s) & GFLOPs \\
\hline
1280$\times$720 & Pixel space (PS)  & 0.00887 & 121.85 \\
                 & Task decoder & 0.00845 & 121.39 \\
1920$\times$1080 & Pixel space (PS)  & 0.00927 & 273.66 \\
                 & Task decoder & 0.00894 & 270.63 \\
\hline
\end{tabular}
\end{table}
\vspace{-5mm}

\subsection{Exportability to Mobile Devices}
We successfully exported the encoder–decoder pipeline to a
Samsung Galaxy S25 Ultra smartphone and executed semantic
segmentation 1072x960 in around 2.5 seconds, demonstrating that the
Shared Neural Space can be applied in resource-constrained
environments, yet with some overhead.




\section{Conclusion}
\label{sec:conc}

We have presented a shared NS, to which different image formats can be mapped, and in which multiple disparate tasks can be performed efficiently with better generalization, resulting in a framework that can enhance applications with multiple modular tasks. 


\vfill\pagebreak



\bibliographystyle{IEEEbib}
\bibliography{refs}

\begin{thebibliography}{10}

\bibitem{donahue2014decaf}
Jeff Donahue, Yangqing Jia, Oriol Vinyals, Judy Hoffman, Ning Zhang, Eric Tzeng, and Trevor Darrell,
\newblock ``Decaf: A deep convolutional activation feature for generic visual recognition,''
\newblock in {\em Proceedings of the 31st International Conference on Machine Learning (ICML)}. 2014, PMLR.

\bibitem{razavian2014offtheshelf}
Ali~Sharif Razavian, Hossein Azizpour, Josephine Sullivan, and Stefan Carlsson,
\newblock ``Cnn features off-the-shelf: An astounding baseline for recognition,''
\newblock in {\em CVPR Workshops}, 2014.

\bibitem{chen2020simclr}
Ting Chen, Simon Kornblith, Mohammad Norouzi, and Geoffrey Hinton,
\newblock ``A simple framework for contrastive learning of visual representations,''
\newblock in {\em Proceedings of the 37th International Conference on Machine Learning (ICML)}. 2020, PMLR.

\bibitem{he2022mae}
Kaiming He, Xinlei Chen, Saining Xie, Yanghao Li, Piotr Doll{\'a}r, and Ross Girshick,
\newblock ``Masked autoencoders are scalable vision learners,''
\newblock in {\em Proceedings of the IEEE/CVF Conference on Computer Vision and Pattern Recognition (CVPR)}, 2022.

\bibitem{radford2021clip}
Alec Radford, Jong~Wook Kim, Chris Hallacy, Aditya Ramesh, Gabriel Goh, Sandhini Agarwal, Girish Sastry, Amanda Askell, Pamela Mishkin, Jack Clark, Gretchen Krueger, and Ilya Sutskever,
\newblock ``Learning transferable visual models from natural language supervision,''
\newblock in {\em Proceedings of the 38th International Conference on Machine Learning (ICML)}. 2021, PMLR.

\bibitem{caruana1997mtl}
Rich Caruana,
\newblock ``Multitask learning,''
\newblock {\em Machine Learning}, vol. 28, pp. 41--75, 1997.

\bibitem{misra2016crossstitch}
Ishan Misra, Abhinav Shrivastava, Abhinav Gupta, and Martial Hebert,
\newblock ``Cross-stitch networks for multi-task learning,''
\newblock in {\em Proceedings of the IEEE Conference on Computer Vision and Pattern Recognition (CVPR)}, 2016.

\bibitem{kendall2018uncertainty}
Alex Kendall, Yarin Gal, and Roberto Cipolla,
\newblock ``Multi-task learning using uncertainty to weigh losses for scene geometry and semantics,''
\newblock in {\em Proceedings of the IEEE Conference on Computer Vision and Pattern Recognition (CVPR)}, 2018.

\bibitem{zamir2018taskonomy}
Amir~Rosenfeld Zamir, Alexander Sax, William Shen, Leonidas Guibas, Jitendra Malik, and Silvio Savarese,
\newblock ``Taskonomy: Disentangling task transfer learning,''
\newblock in {\em Proceedings of the IEEE Conference on Computer Vision and Pattern Recognition (CVPR)}, 2018.

\bibitem{kokkinos2017ubernet}
Iasonas Kokkinos,
\newblock ``Ubernet: Training a universal convolutional neural network for low-, mid-, and high-level vision using diverse datasets and limited memory,''
\newblock in {\em Proceedings of the IEEE Conference on Computer Vision and Pattern Recognition (CVPR)}, 2017.

\bibitem{vandenhende2020mtinet}
Simon Vandenhende, Stamatios Georgoulis, and Luc Van~Gool,
\newblock ``Mti-net: Multi-scale task interaction networks for multi-task learning,''
\newblock in {\em Proceedings of the European Conference on Computer Vision (ECCV)}, 2020.

\bibitem{tsai2018adaptsegnet}
Yi-Hsuan Tsai, Wei-Chih Hung, Samuel Schulter, Kihyuk Sohn, Ming-Hsuan Yang, and Manmohan Chandraker,
\newblock ``Learning to adapt structured output space for semantic segmentation,''
\newblock in {\em Proceedings of the IEEE Conference on Computer Vision and Pattern Recognition (CVPR)}, 2018.

\bibitem{hoffman2018cycada}
Judy Hoffman, Eric Tzeng, Taesung Park, Jun-Yan Zhu, Phillip Isola, Kate Saenko, Alexei~A. Efros, and Trevor Darrell,
\newblock ``Cycada: Cycle-consistent adversarial domain adaptation,''
\newblock in {\em Proceedings of the 35th International Conference on Machine Learning (ICML)}. 2018, PMLR.

\bibitem{zhou2021mixstyle}
Kaiyang Zhou, Yongxin Yang, Yu~Qiao, and Tao Xiang,
\newblock ``Domain generalization with mixstyle,''
\newblock in {\em International Conference on Learning Representations (ICLR)}, 2021.

\bibitem{cha2021swad}
Jang-Hyun Cha, Keonwoo Choi, Dongyoon Lee, and Sanghyuk Chun,
\newblock ``Swad: Domain generalization by seeking flat minima,''
\newblock in {\em Advances in Neural Information Processing Systems (NeurIPS)}, 2021.

\bibitem{Li_2025_CVPR}
Jing Li, Chengyu Wang, Hamid~R. Sheikh, and SeokJun Lee,
\newblock ``Cdvs: Compressed domain on device memory efficient 8k video slowmo,''
\newblock in {\em Proceedings of the IEEE/CVF Conference on Computer Vision and Pattern Recognition (CVPR) Workshops}, June 2025, pp. 1947--1953.

\bibitem{kouzelis2025eqvae}
Theodoros Kouzelis, Ioannis Kakogeorgiou, Spyros Gidaris, and Nikos Komodakis,
\newblock ``{EQ}-{VAE}: Equivariance regularized latent space for improved generative image modeling,''
\newblock in {\em Forty-second International Conference on Machine Learning}, 2025.

\bibitem{xing2022residualswintransformerchannel}
Wenzhu Xing and Karen Egiazarian,
\newblock ``Residual swin transformer channel attention network for image demosaicing,'' 2022.

\bibitem{zhang2017beyond}
Kai Zhang, Wangmeng Zuo, Yunjin Chen, Deyu Meng, and Lei Zhang,
\newblock ``Beyond a gaussian denoiser: Residual learning of deep cnn for image denoising,''
\newblock {\em IEEE transactions on image processing}, vol. 26, no. 7, pp. 3142--3155, 2017.

\end{thebibliography}

\end{document}